\documentclass[letterpaper, 10 pt, conference]{ieeeconf}  % Comment this line out if you need a4paper

\IEEEoverridecommandlockouts                              % This command is only needed if 
\usepackage{amssymb,amsmath,amsfonts,mathrsfs}
\usepackage{graphicx}
\usepackage[table,xcdraw,dvipsnames]{xcolor}
\usepackage{tikz}
\usepackage{url}

\usepackage[utf8]{inputenc}

\usepackage[labelformat=simple]{subcaption}

\usepackage{colortbl}
\usepackage{booktabs}
\usepackage{array}
\usepackage{tabularx}

\usepackage[flushleft]{threeparttable}
\usepackage{multirow}

\let\labelindent\relax
\usepackage{enumitem}

\newcolumntype{L}[1]{>{\raggedright\let\newline\\\arraybackslash\hspace{0pt}}m{#1}}

\makeatletter
\DeclareRobustCommand\onedot{\futurelet\@let@token\@onedot}
\def\@onedot{\ifx\@let@token.\else.\null\fi\xspace}

\usepackage[T1]{fontenc}  % For correct {}s in \texttt
\usepackage[b]{esvect}    % For \vv

\makeatletter
\newlength\xvec@height%
\newlength\xvec@depth%
\newlength\xvec@width%
\newcommand{\xvec}[2][]{%
  \ifmmode%
    \settoheight{\xvec@height}{$#2$}%
    \settodepth{\xvec@depth}{$#2$}%
    \settowidth{\xvec@width}{$#2$}%
  \else%
    \settoheight{\xvec@height}{#2}%
    \settodepth{\xvec@depth}{#2}%
    \settowidth{\xvec@width}{#2}%
  \fi%
  \def\xvec@arg{#1}%
  \def\xvec@dd{:}%
  \def\xvec@d{.}%
  \raisebox{.2ex}{\raisebox{\xvec@height}{\rlap{%
    \kern.05em%  (Because left edge of drawing is at .05em)
    \begin{tikzpicture}[scale=1]
    \pgfsetroundcap
    \draw (.05em,0)--(\xvec@width-.05em,0);
    \draw (\xvec@width-.05em,0)--(\xvec@width-.15em, .075em);
    \draw (\xvec@width-.05em,0)--(\xvec@width-.15em,-.075em);
    \ifx\xvec@arg\xvec@d%
      \fill(\xvec@width*.45,.5ex) circle (.5pt);%
    \else\ifx\xvec@arg\xvec@dd%
      \fill(\xvec@width*.30,.5ex) circle (.5pt);%
      \fill(\xvec@width*.65,.5ex) circle (.5pt);%
    \fi\fi%
    \end{tikzpicture}%
  }}}%
  #2%
}
\makeatother

\makeatletter
\renewcommand*\env@matrix[1][\arraystretch]{%
  \edef\arraystretch{#1}%
  \hskip -\arraycolsep
  \let\@ifnextchar\new@ifnextchar
  \array{*\c@MaxMatrixCols c}}
\makeatother

\newcommand{\unit}[1]{\left[ #1 \right]}

\definecolor{commentcolor}{gray}{0.5}
\usepackage{algorithm}
\usepackage{algpseudocode}
\algrenewcommand\algorithmicindent{1.0em}%

\algnewcommand{\LineComment}[1]{\State \textcolor{commentcolor}{\(\triangleright\) #1}}
\algnewcommand{\NewComment}[1]{\textcolor{commentcolor}{\(\triangleright\) #1}}
\algnewcommand{\To}{\textbf{to}}
\algnewcommand{\Break}{\textbf{break}}
\algnewcommand{\Continue}{\textbf{continue}}
\algnewcommand{\IIf}[1]{\State\algorithmicif\ #1\ \algorithmicthen}
\algnewcommand{\EndIIf}{\unskip}
\algnewcommand{\var}[1]{\textit{#1}}
\algnewcommand{\func}[1]{\textsc{#1}}

\def\publishedurl{\centering \textbf{{\color{red}{Presented at 2nd Workshop on Representing and Manipulating Deformable Objects}}\\
{\color{red}{The 2022 International Conference on Robotics and Automation (ICRA), May 23-27, 2022}}\\
\color{blue}{\url{https://deformable-workshop.github.io/icra2022/}}}}

\title{\LARGE \bf
Detection and Physical Interaction with Deformable Linear Objects}

\author{Azarakhsh Keipour$^{1}$, 
Mohammadreza Mousaei$^{2}$,
Maryam Bandari$^{3}$, 
Stefan Schaal$^{4}$
and Sebastian Scherer$^{5}$ % <-this % stops a space
\thanks{$^{1}$ Robotics Institute, Carnegie Mellon University, Pittsburgh, PA
        {\tt\small keipour@gmail.com}}%
\thanks{$^{2,5}$ Robotics Institute, Carnegie Mellon University, Pittsburgh, PA {\tt\small [mmousaei, basti]@cmu.edu}}%
\thanks{$^{3,4}$ Intrinsic, Mountain View, CA {\tt\small [maryamb, sschaal]@intrinsic.ai}}%
}

\begin{document}

\maketitle
\thispagestyle{empty}
\pagestyle{empty}

%%%%%%%%%%%%%%%%%%%%%%%%%%%%%%%%%%%%%%%%%%%%%%%%%%%%%%%%%%%%%%%%%%%%%%%%%%%%%%%%
\begin{abstract}

Deformable linear objects (e.g., cables, ropes, and threads) commonly appear in our everyday lives. However, perception of these objects and the study of physical interaction with them is still a growing area. There have already been successful methods to model and track deformable linear objects. However, the number of methods that can automatically extract the initial conditions in non-trivial situations for these methods has been limited, and they have been introduced to the community only recently. On the other hand, while physical interaction with these objects has been done with ground manipulators, there have not been any studies on physical interaction and manipulation of the deformable linear object with aerial robots.

This workshop describes our recent work on detecting deformable linear objects, which uses the segmentation output of the existing methods to provide the initialization required by the tracking methods automatically. It works with crossings and can fill the gaps and occlusions in the segmentation and output the model desirable for physical interaction and simulation. Then we present our work on using the method for tasks such as routing and manipulation with the ground and aerial robots. We discuss our feasibility analysis on extending the physical interaction with these objects to aerial manipulation applications.

\end{abstract}

%%%%%%%%%%%%%%%%%%%%%%%%%%%%%%%%%%%%%%%%%%%%%%%%%%%%%%%%%%%%%%%%%%%%%%%%%%%%%%%%

\section{Introduction} \label{sec:intro}

Deformable One-dimensional Objects (DOOs) or Deformable Linear Objects (DLOs) are a class of objects that includes ropes, cables, threads, sutures, and wires. A crucial part of achieving full autonomy for physical interaction with DLOs is perception. Many applications require complete knowledge of the object's initial state as a model that allows simulation and the computation of its dynamics.

Many researchers in the medical and industrial community have proposed methods to extract DLOs from images~\cite{KRISSIAN2000130, 28493, Wang2021, Noble2011, 32226}. These methods mainly provide the region in the image containing the DLO and are not directly suitable for autonomous physical interaction. On the other hand, various algorithms have been proposed to track a DLO across the video frames~\cite{5432191, 5980431, 15711, wang2020tracking, rastegarpanah, 6630714, 8560497, af66d77e53304e6bbb64049bb46193cb, 8206058}. These methods have been used for physical interaction and manipulation with ground robots, and while some of these methods can detect the initial DLO state in trivial conditions (e.g., a straight rope in camera view), others require even a simple DLO configuration to be provided to them a~priori. On the other hand, the aerial robotics community has extensively worked on the segmentation of cables and wires for visual inspection and obstacle avoidance purposes. These methods can effectively segment out the power lines, bridge cables, and other near-straight deformable objects~\cite{8206190, 8577142, 7532456, Dai2020, 7279641, 7181891, PAGNANO2013234, 6322366}. However, none of these approaches can handle true deformations, making these approaches unsuitable for physical interaction tasks.

While providing the initial DLO state in the lab settings is possible, it is not generally provided in real-world applications. On the other hand, the existing learning-based detection methods are labor-intensive to train for each new DLO and new setting and are hard to generalize to less-certain conditions of the real-world~\cite{8972568}.

\begin{figure}[!t]
\centering
    \begin{subfigure}[b]{0.125\textwidth}
        \includegraphics[width=\textwidth, height=2.1cm]{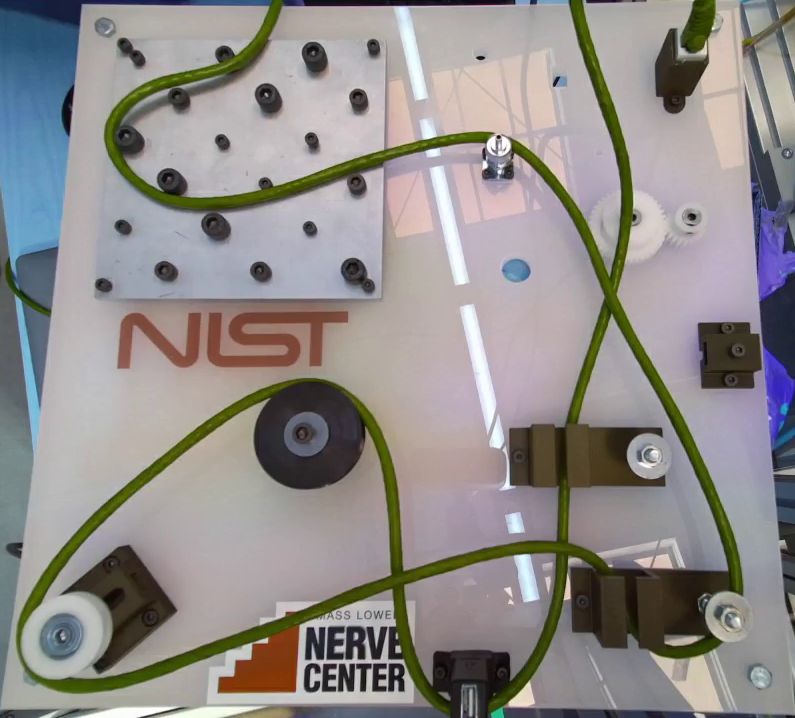}
    \end{subfigure}
    \hfill
    \begin{subfigure}[b]{0.125\textwidth}
        \includegraphics[width=\textwidth, height=2.1cm]{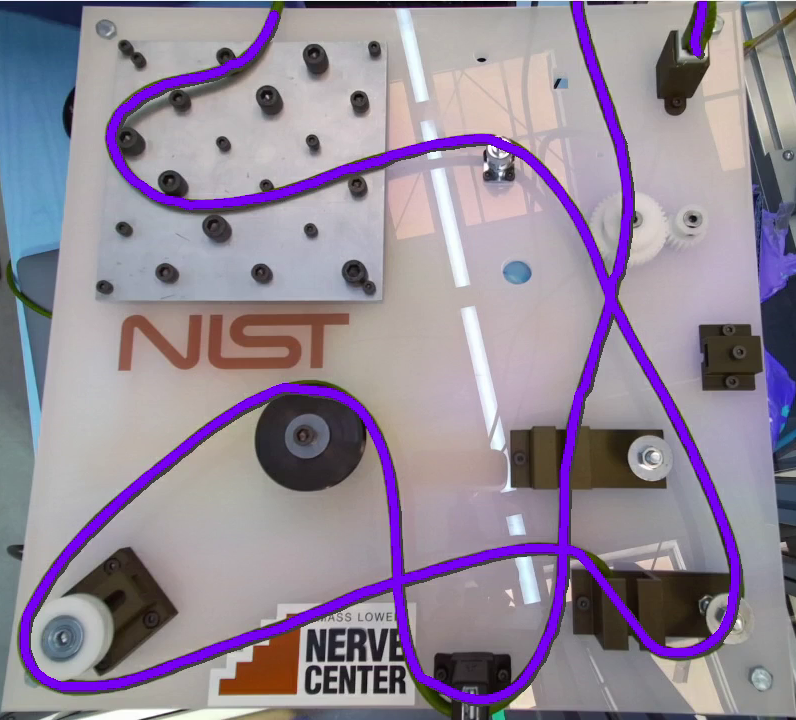}
    \end{subfigure}
    ~
    \begin{subfigure}[b]{0.10\textwidth}
        \includegraphics[width=\textwidth, height=2.1cm]{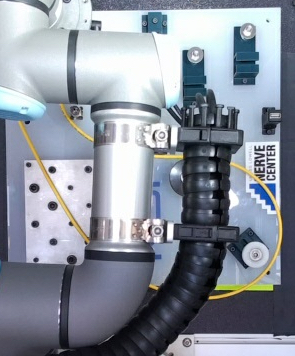}
    \end{subfigure}
    \hfill
    \begin{subfigure}[b]{0.10\textwidth}
        \includegraphics[width=\textwidth, height=2.1cm]{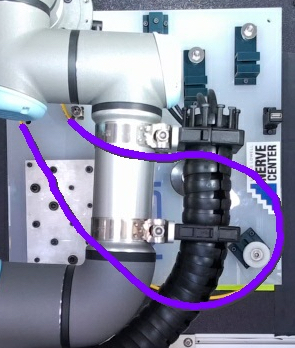}
    \end{subfigure}
    
    \medskip
    
    \begin{subfigure}[b]{0.48\textwidth}
        \includegraphics[width=0.20\textwidth, height=2.2cm]{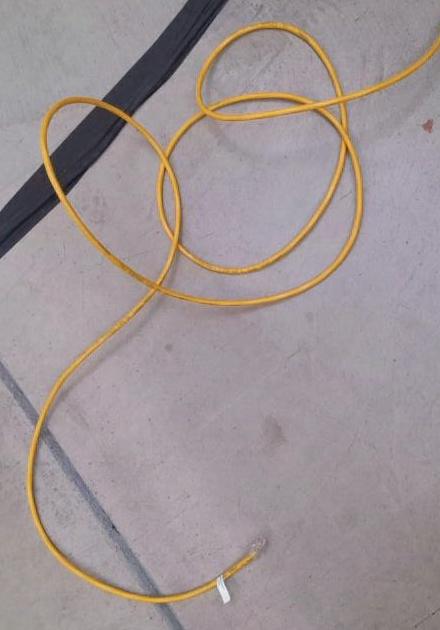}
        \hfill
        \includegraphics[width=0.20\textwidth, height=2.2cm]{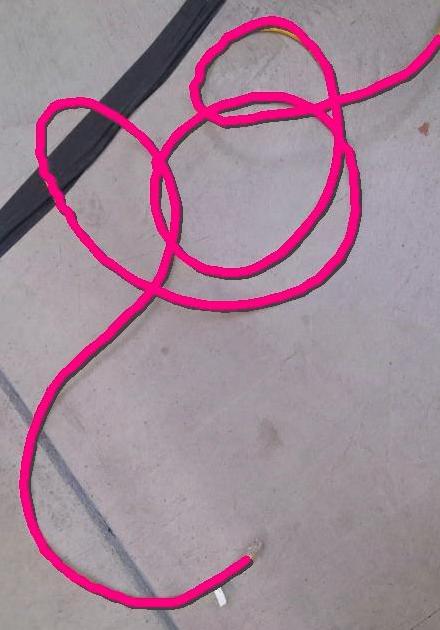}
        ~
        \includegraphics[width=0.29\textwidth, height=2.2cm]{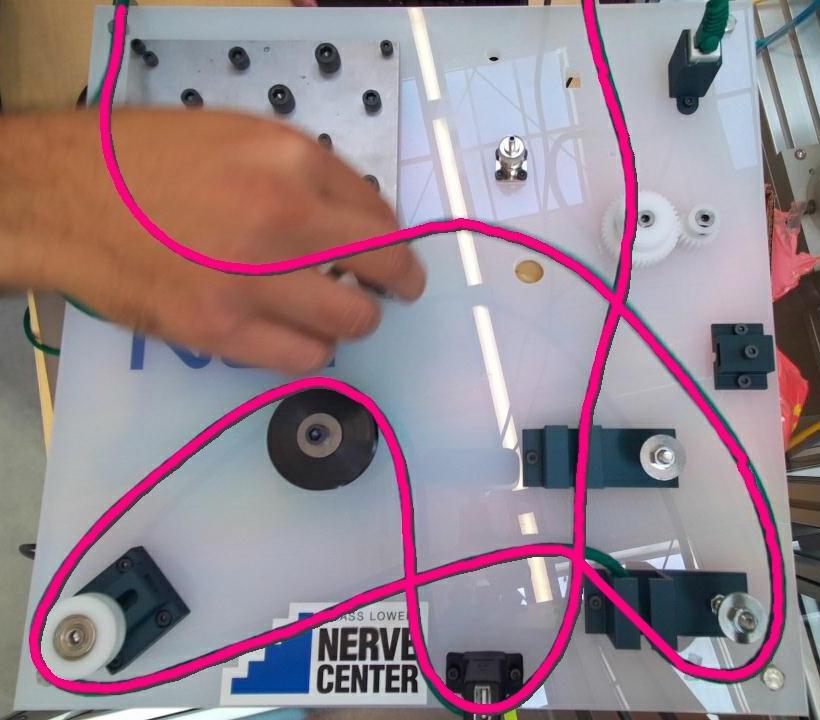}
        ~
        \includegraphics[width=0.22\textwidth, height=2.2cm]{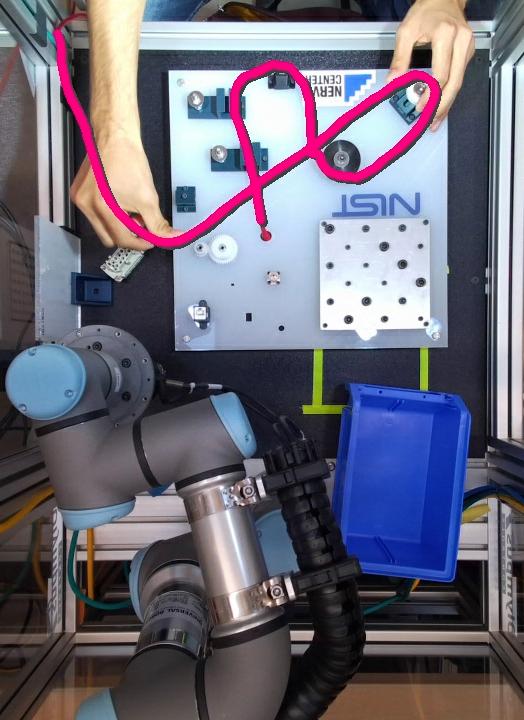}
    \end{subfigure}
    \caption{The result of DLO detection (overlaid with purple and pink) on inputs with occlusions and crossings~\cite{keipour2022ral}.}
\label{fig:results-detection}
\end{figure}

This workshop briefly reviews our approach from~\cite{keipour2022ral} to detect the initial conditions of a deformable linear object in more complex scenarios, with gaps, occlusions, and DLO crossings. We further analyze our tested applications of this method in ground manipulation and routing and study the feasibility of tasks involving the physical interaction of DLOs for aerial robots. 

\section{The Method} \label{sec:method}

Our detection method outputs the DLO state as a chain of fixed-sized segments connected by passive spherical joints, commonly used for manipulation and dynamic simulation. The method has six steps: segmentation, skeletonization, contour extraction, DLO fitting, pruning, and merging. The first three steps can use a combination of existing approaches, while the rest of the steps are specific to this method. The algorithm is briefly described in this section but is explained in more detail in~\cite{keipour2022ral}. 

\subsection{Segmentation, Skeletonization and Contour Extraction}

Segmentation filters the image data to extract the DLO portions and exclude all other data. Many model-based and learning-based segmentation methods have been proposed in both medical and industrial robotics communities~\cite{KRISSIAN2000130, 28493, Wang2021, Noble2011, 32226, 10605, BFb0056195, 72a4e1c53c}. This step should eliminate all the unrelated data, even if it removes some DLO data.

Skeletonization transforms each segmented connected component into a set of connected pixels with single-pixel width. It is commonly used in many applications~\cite{SAHA20173, keipour2013omnifont}. Our algorithm requires the skeleton of each connected component to remain connected and only one branch to be returned per actual branch. We used a well-known morphological thinning method for skeletonization~\cite{1164959}. 

Extracting contours is also a standard step in many applications~\cite{Gong2018, keipour2021ral}. The contour extraction methods can result in several contours per branch and some with multiple branches. Our DLO detection method can handle these issues, and many of the existing contour extraction methods can be used with it. We used Suzuki and Abe's method~\cite{suzuki1985topological} to extract contours.

\begin{figure}[!t]
\centering
    \begin{subfigure}[b]{0.23\linewidth}
        \includegraphics[width=\textwidth, height=1.8cm]{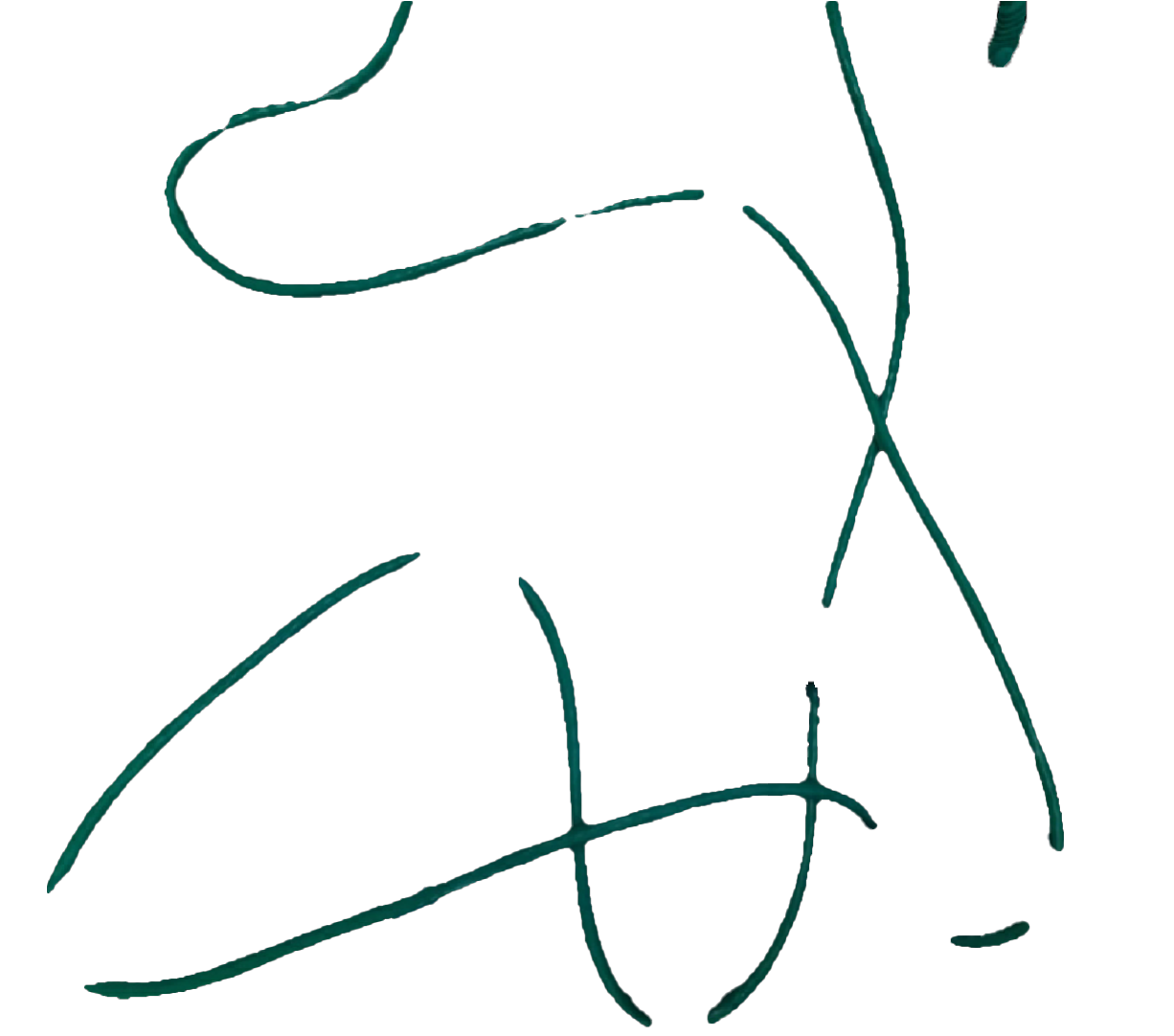}
        \caption{~}
        \label{fig:segmented-board}
    \end{subfigure}
    \hfill
    \begin{subfigure}[b]{0.21\linewidth}
        \includegraphics[width=\textwidth, height=1.8cm]{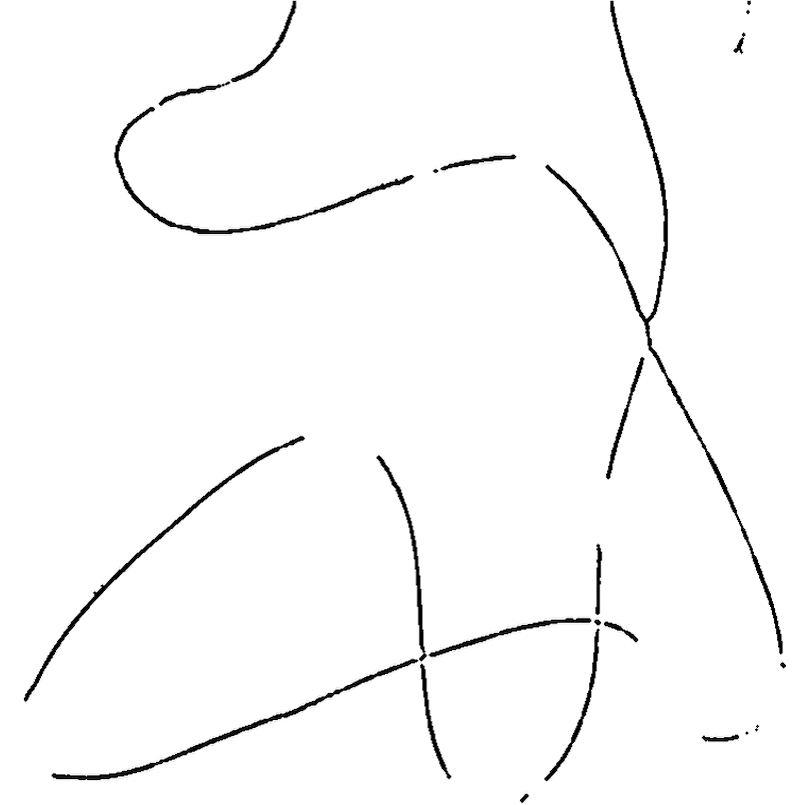}
        \caption{~}
        \label{fig:skeletonization}
    \end{subfigure}
    \hfill
    \begin{subfigure}[b]{0.21\linewidth}
        \includegraphics[width=\textwidth, height=1.8cm]{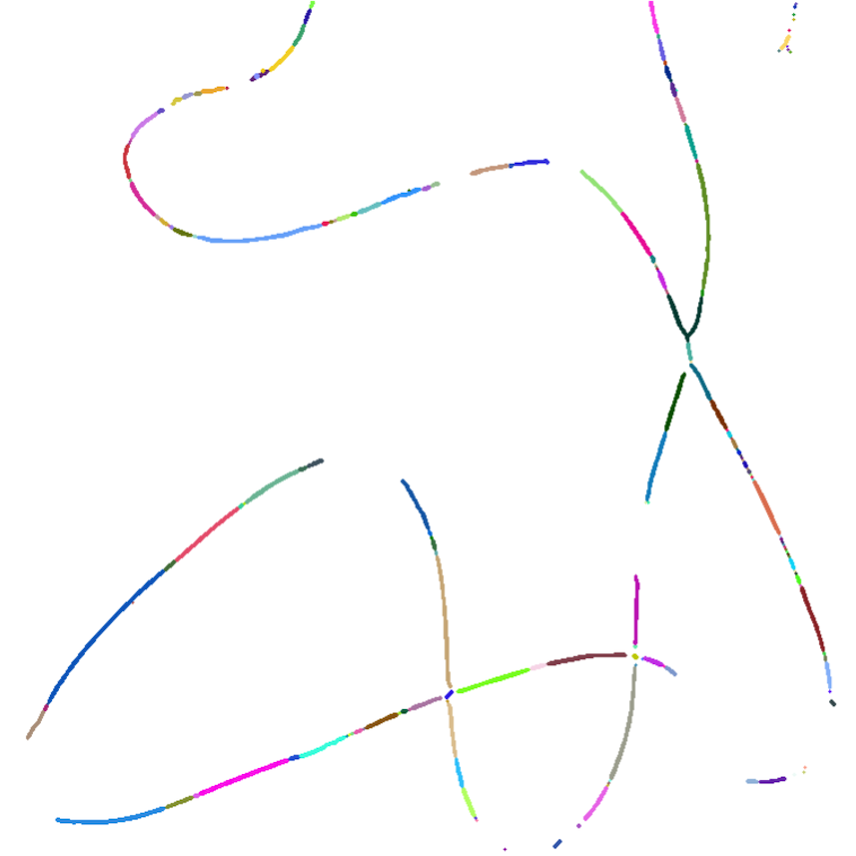}
        \caption{~}
        \label{fig:contour-extraction}
    \end{subfigure}
    \hfill
    \begin{subfigure}[b]{0.21\linewidth}
        \includegraphics[width=\textwidth, height=1.8cm]{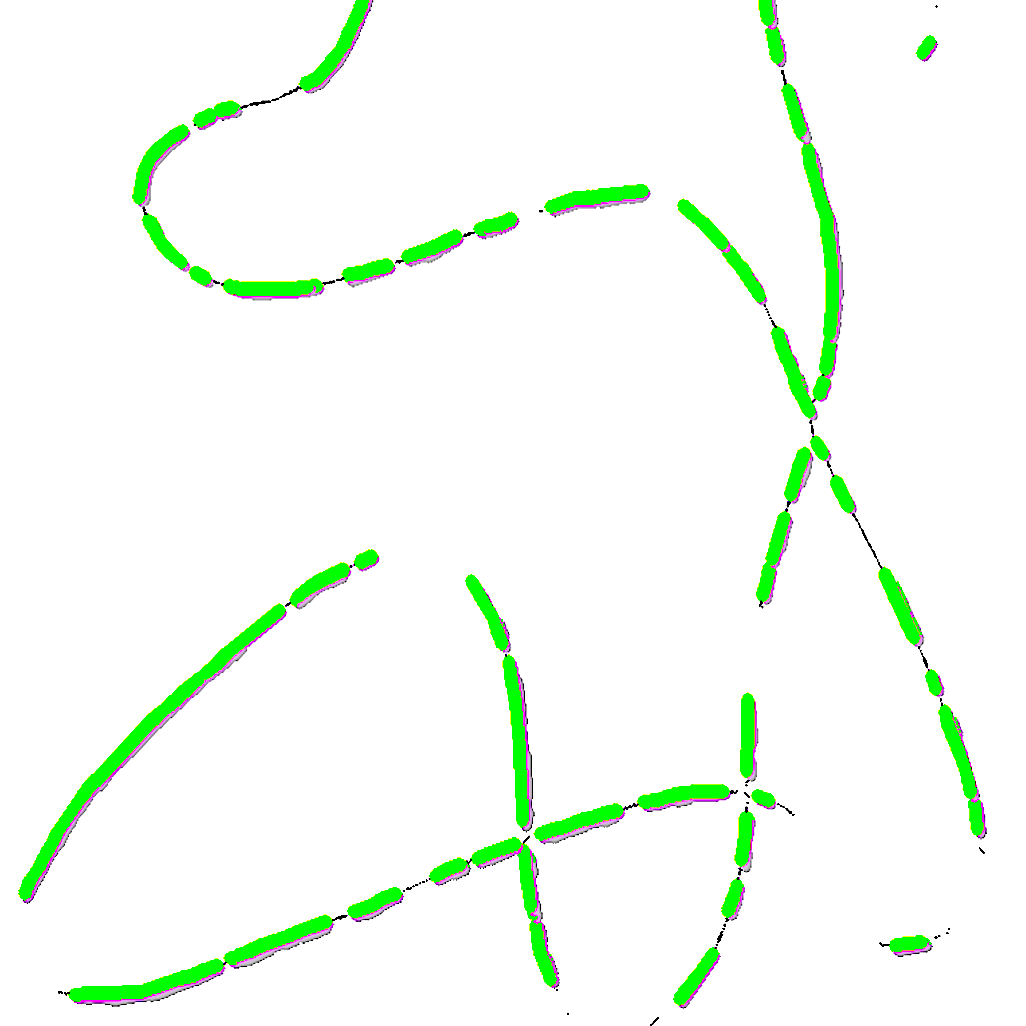}
        \caption{~}
        \label{fig:fitting-pruning}
    \end{subfigure}
    \caption{The first four steps of DLO detection on an image frame from Figure~\ref{fig:results-detection}~\cite{keipour2022ral}. (a) Segmentation. (b) Skeletonization. (c) Contour extraction. (d) DLO Fitting and Pruning.}
    \label{fig:first-four-steps}
\end{figure}

Figures~\ref{fig:first-four-steps}(a-c) present the results of the first three steps on an example camera frame.

\subsection{DLO Fitting and Pruning} \label{sec:fitting-pruning}

Each contour can be a single branch, or it may contain multiple branches. The pixel sequence for a contour returned by a typical contour extraction method starts from one of the branch tips, goes around the skeleton component, and ends with a sharp turn back at the start point. 

Let us define the length of each DLO segment (i.e., the fixed cylinder) as $l_s$. The DLO fitting step initializes an empty DLO chain, starts from the first pixel in the contour, and traverses over the pixels until the distance from that pixel is $l_s$. It creates a new cylindrical segment, adds it to the chain, and continues traversing from the end of the new cylinder again. Every time it goes over a branch tip, it records the current DLO chain and initializes a new one. This continues until traversing reaches the last point in the contour sequence.

This method returns multiple overlapping DLO chains for each part of the object, and overlapping segments are pruned to simplify the further steps by assuming that no segments overlap. Figure~\ref{fig:fitting-pruning}(d) shows the result of fitting and pruning.

% ----------------------------------------------

\subsection{Merging}

With the collection of DLO chains, they are iteratively merged to fill the gaps and form a single object. Each iteration connects two chains, and the process is repeated until there is only one chain left. With two chains each having two ends, there are four cost values for connecting the two ends for any two chains. The lowest among the four values is the cost of connecting the two chains, and the chain pair with the lowest cost is connected at each iteration.

\begin{figure}[!t]
    \centering
        \begin{subfigure}[b]{0.1\textwidth}
            \includegraphics[width=\textwidth]{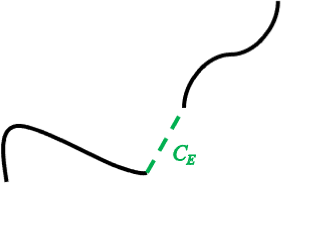}
            \caption{~}
            \label{fig:merge-costs-e}
        \end{subfigure}
        \hfill
        \begin{subfigure}[b]{0.1\textwidth}
            \includegraphics[width=\textwidth]{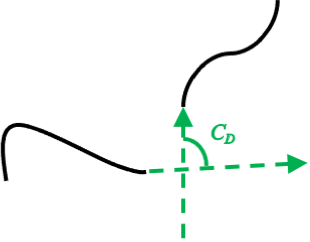}
            \caption{~}
            \label{fig:merge-costs-d}
        \end{subfigure}
        \hfill
        \begin{subfigure}[b]{0.2\textwidth}
            \includegraphics[width=\textwidth]{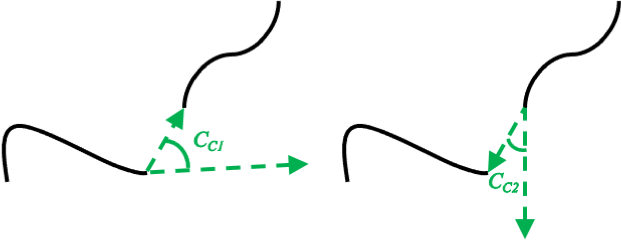}
            \caption{~}
            \label{fig:merge-costs-c}
        \end{subfigure}
    \caption{Partial merging costs for two chain ends~\cite{keipour2022ral}. (a) Euclidean cost. (b) Direction cost. (c) Curvature costs.}
    \label{fig:merge-costs}
    \end{figure}

Three separate partial costs are used for computing the total connection cost: the Euclidean distance of the two chain ends (Figure~\ref{fig:merge-costs-e}), the difference in their directions (Figure~\ref{fig:merge-costs-d}), and how much curvature is needed to connect them (Figure~\ref{fig:merge-costs-c}).

% ----------------------------------------------

\begin{figure}[!t]
\centering
    \begin{subfigure}[b]{0.09\textwidth}
        \includegraphics[width=\textwidth, height=1cm]{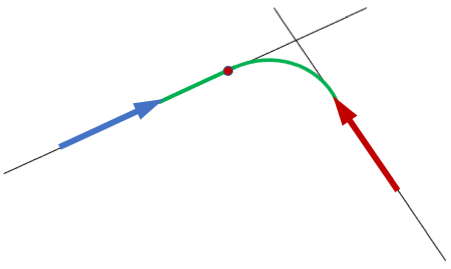}
        \caption{~}
    \end{subfigure}
    \hfill
    \begin{subfigure}[b]{0.09\textwidth}
        \includegraphics[width=\textwidth, height=1cm]{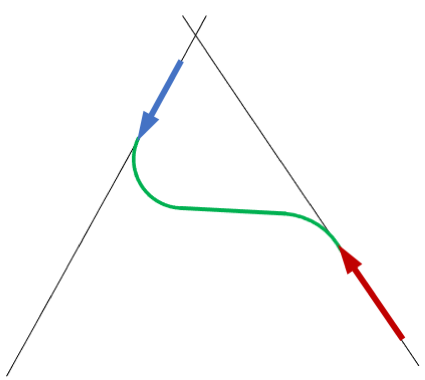}
        \caption{~}
    \end{subfigure}
    \hfill
    \begin{subfigure}[b]{0.18\textwidth}
        \includegraphics[width=\textwidth, height=1cm]{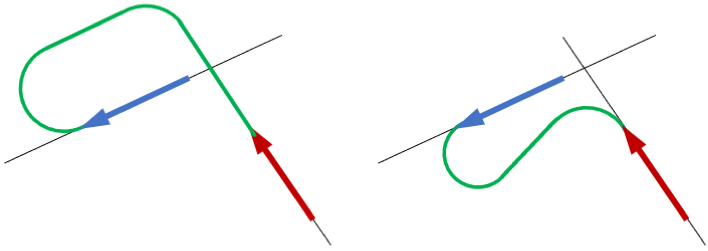}
        \caption{~}
    \end{subfigure}
\caption{Illustration of suggested solutions for merging scenarios~\cite{keipour2022ral}. The arrows represent the DLO chain ends, with the arrow sides representing the end points.}
\label{fig:merge-cases-fill}
\end{figure}

The gap between the two chains should be filled to follow the expected curve of the deformable object. While any deformable object can take almost any curve, it is possible to have an educated \textit{guess}. We calculate the "natural" curvature, which connects the desired ends of the two chains. The "natural" curvature initially follows the direction of the last segments of the chains and then follows a curve with the largest constant turn rate possible. Our suggested solutions for the different possible cases are shown in Figure~\ref{fig:merge-cases-fill}.

\section{Experiments and Results} \label{sec:tests}

Figure~\ref{fig:results-detection} shows the detection results on some input frames. The average detection time per frame across all the sequences is 0.537 seconds on a system with Intel® Core™ i9-10885H CPU and 64 GB DDR4 RAM and a sub-optimal implementation in Python~3. We tested the method on 7~video sequences with 4,230~frames of size 1280$\times$720. Table~\ref{tbl:results} shows the quantitative results for the algorithm's accuracy on the whole cable in an image, for the occlusions filled, and for the merges performed. For a more detailed explanation of the results, please see~\cite{keipour2022ral}.

\begin{table}[!t]
\centering
\caption{Detection results on 7 video sequences~\cite{keipour2022ral}.}
\label{tbl:results}
\begin{tabular}{|c|c|c|c|c|c|c|c|c|c|}
\hline
\rowcolor[HTML]{EFEFEF} 
~& Total & Correct & Incorrect & Accuracy \\ \hline
Frames         &  4,230 & 3,542 & 688 & 83.7\% \\ \hline
Occlusions & 26,456 & 23,991 & 2,465 & 90.7\% \\ \hline
Merges & 583,743 & 581,130 & 2,613 & 99.6\%\\ \hline
\end{tabular}
\end{table}

\section{Feasibility Analysis for Manipulation}

\begin{figure}[!t]
\centering
    \includegraphics[width=0.24\linewidth, height=2cm]{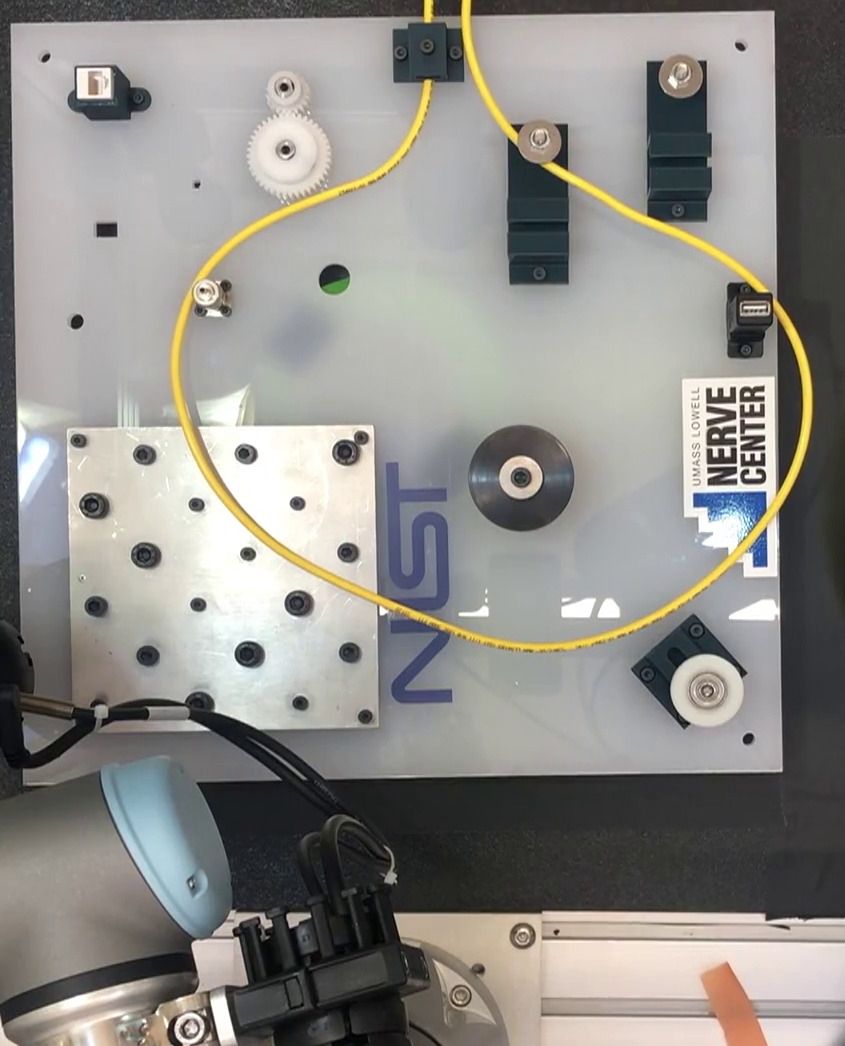}
    \hfill    
    \includegraphics[width=0.24\linewidth, height=2cm]{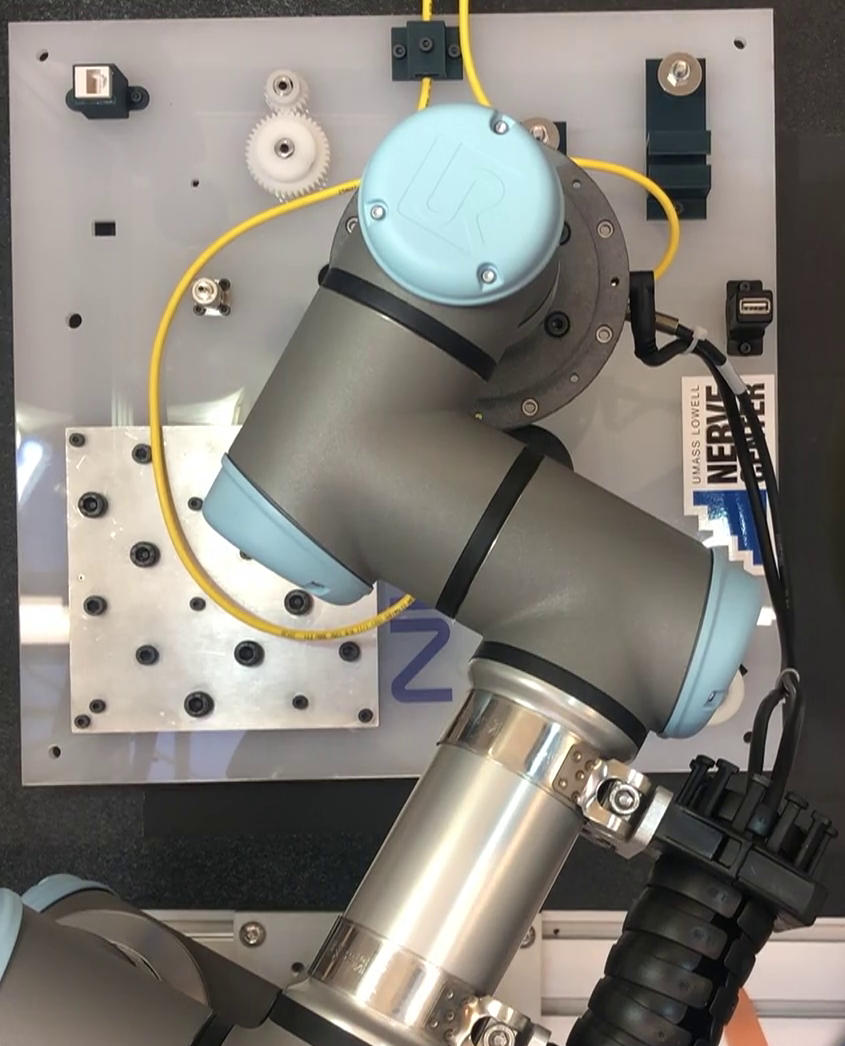}
    \hfill
    \includegraphics[width=0.24\linewidth, height=2cm]{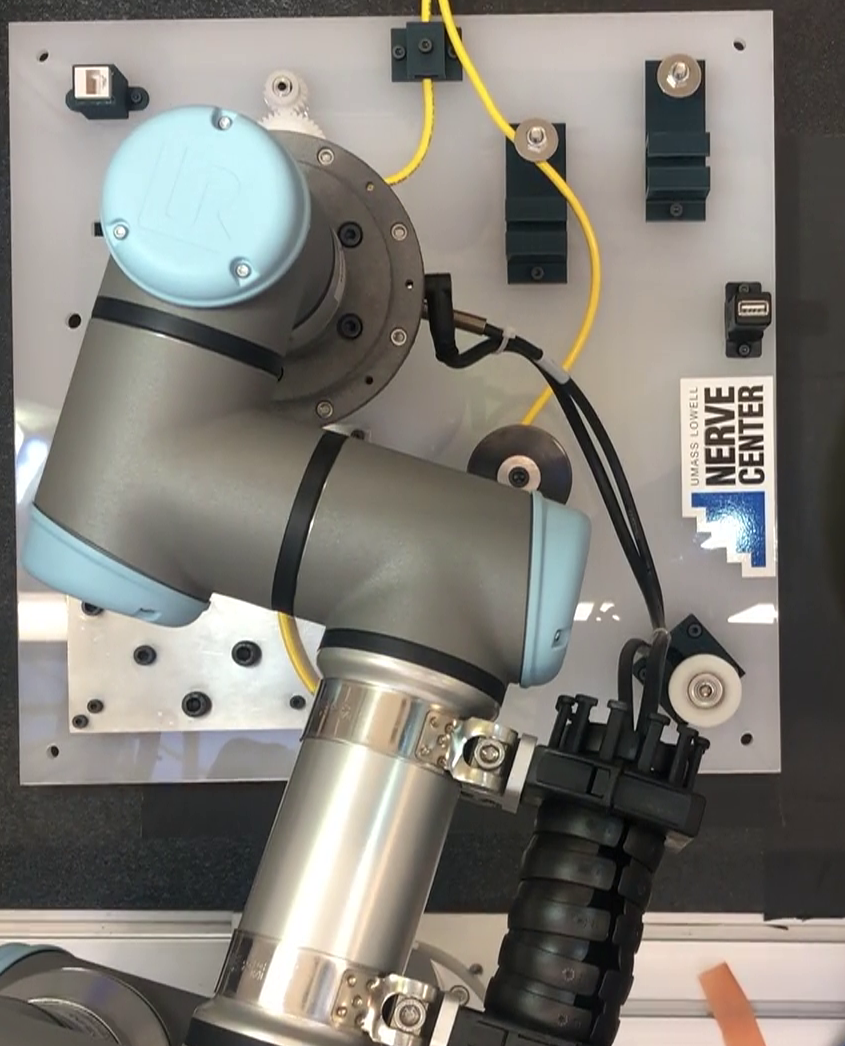}
    \hfill    
    \includegraphics[width=0.24\linewidth, height=2cm]{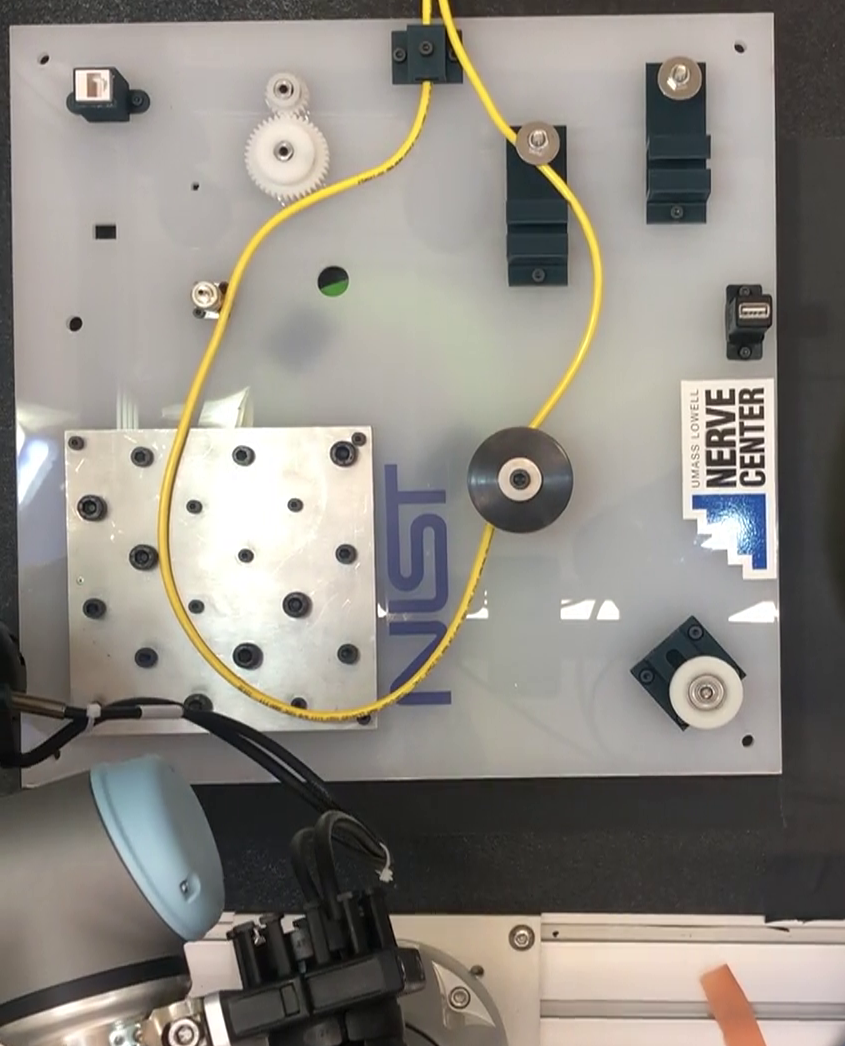}
    \caption{Cable routing and manipulation using a UR3 robotic arm~\cite{keipour2022iros}.}
    \label{fig:results-manipulation}
\end{figure}

To test the feasibility of the detection method for fully-automated physical interaction, we have used it in a cable routing and manipulation application with a ground UR3 manipulator arm~\cite{keipour2022iros}. Figure~\ref{fig:results-manipulation} shows screenshots from one of the tests.

\begin{figure}[!t]
    \centering
        \begin{subfigure}[b]{0.36\linewidth}
            \includegraphics[width=\textwidth]{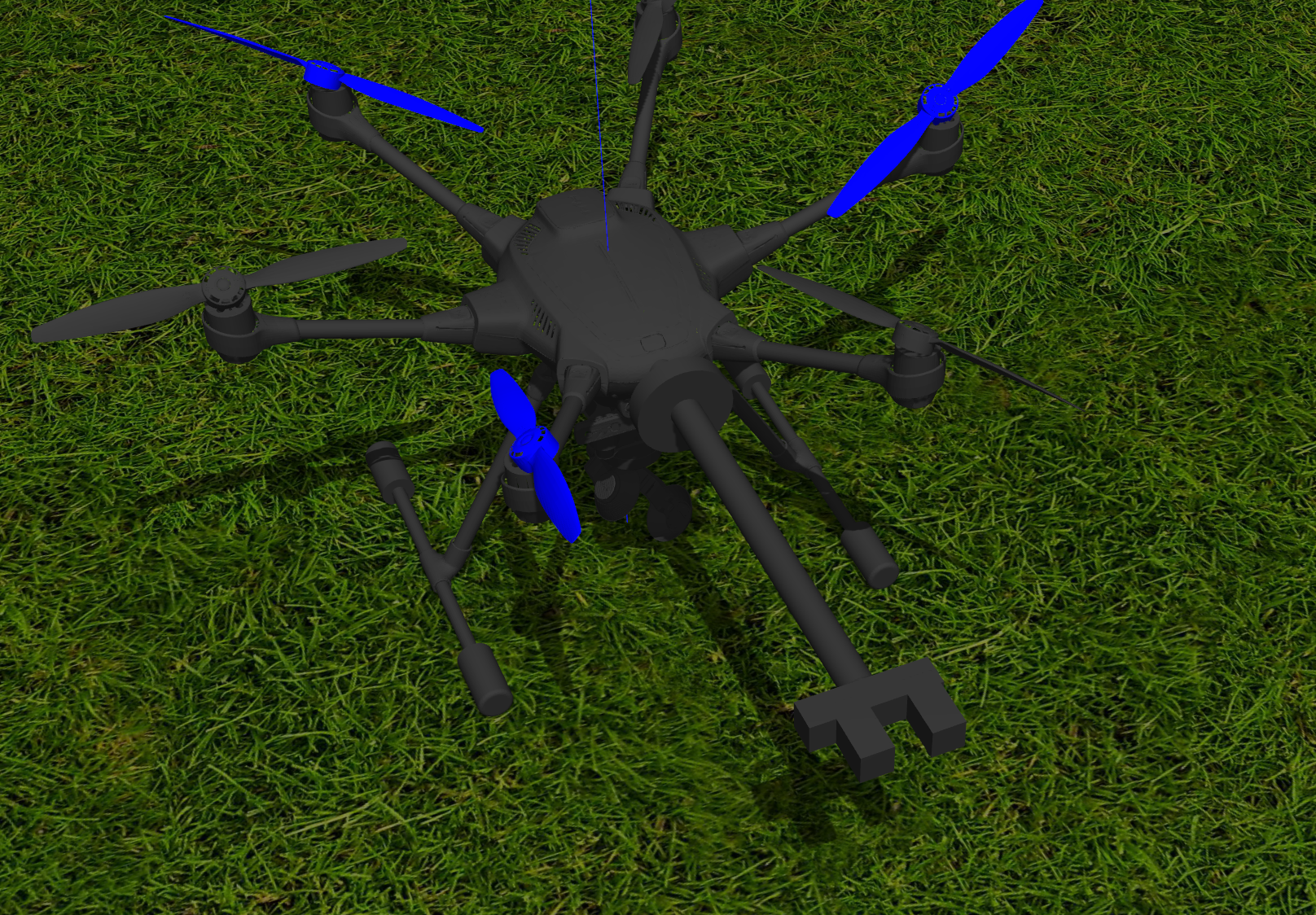}
            \caption{~}
            \label{fig:results-uav-gazebo}
        \end{subfigure}
        \hfill
        \begin{subfigure}[b]{0.61\linewidth}
            \includegraphics[width=0.59\textwidth]{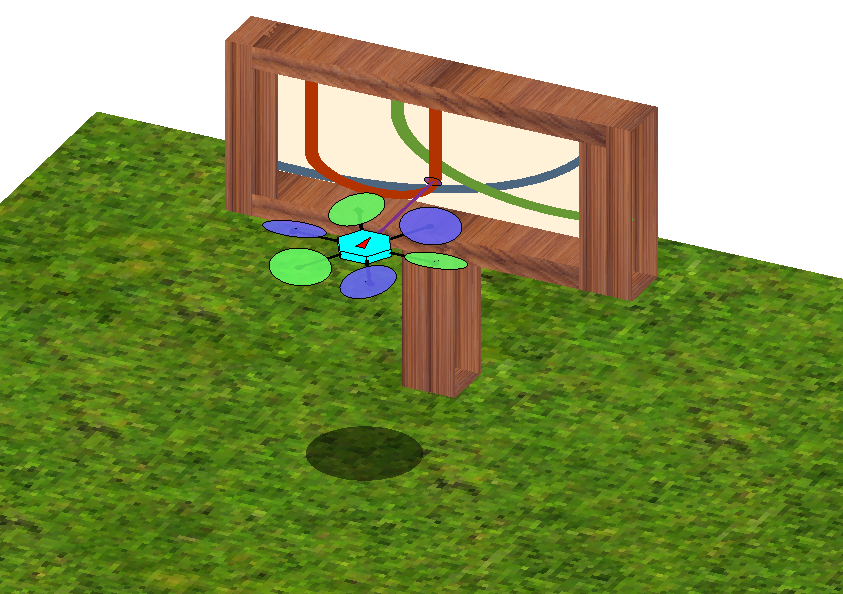}
            \hfill
            \includegraphics[width=0.38\textwidth]{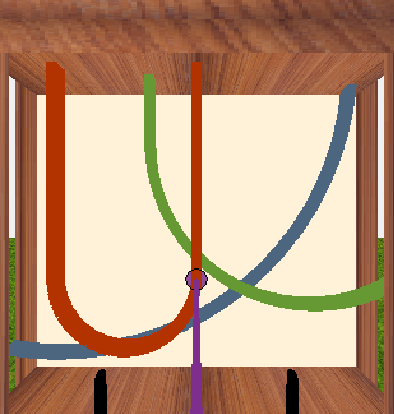}
            \caption{~}
            \label{fig:results-uav-matlab}
        \end{subfigure}
    \caption{Feasibility tests of DLO manipulation using aerial robots in simulation. (a) Gazebo model of our robot with a gripper for cable manipulation. (b) MATLAB cable grasping tests in ARCAD simulator~\cite{Keipour:2023:scitech:simulator, Keipour:2023:icra-workshop:simulator}.}
    \label{fig:results-uav}
    \end{figure}

In general, ground manipulators have higher precision compared to aerial robots. For aerial DLO manipulation (for example, for maintenance and cable manipulation at the top of utility poles), a significant issue is the precision of the end-effector in grasping the DLOs detected in the camera. We tested the feasibility of manipulating the detected cable in Gazebo and MATLAB for a fully-actuated hexarotor with tilted arms controlled with the system introduced in~\cite{keipour2020integration}. We measured the position error for grasping a specific point on the cable~\cite{Keipour-2022-131692}. Figure~\ref{fig:results-uav} shows our setup for testing the feasibility of the task. The error from the MATLAB simulator is near zero. The Gazebo simulator tends to give more realistic results, so we only report the Gazebo experiments. 

\begin{figure}[!t]
        \centering
        \begin{subfigure}[b]{0.48\linewidth}
            \includegraphics[width=\textwidth, height=3cm]{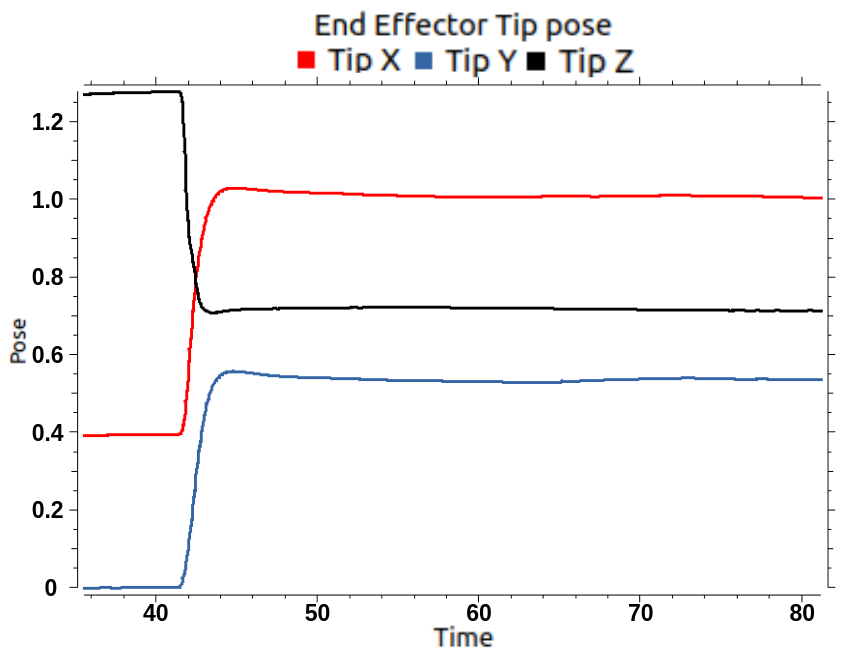}
            \caption{~}
            \label{fig:results-uav-end-effector}
        \end{subfigure}
        \begin{subfigure}[b]{0.48\linewidth}
            \includegraphics[width=\textwidth, height=3cm]{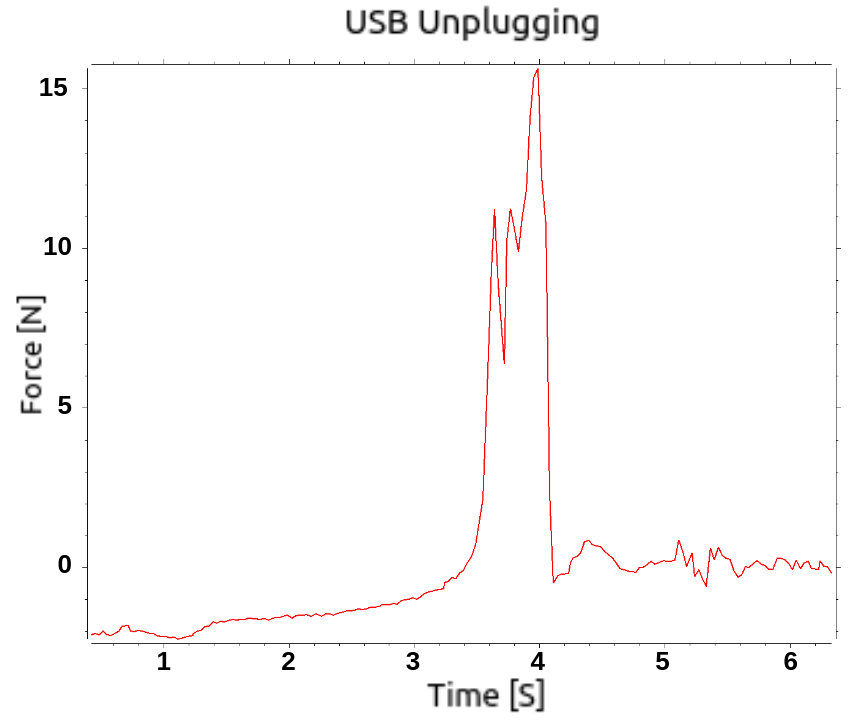}
            \caption{~}
            \label{fig:results-usb-unplug-plot}
        \end{subfigure}
        \caption{Feasibility tests of aerial DLO manipulation in Gazebo. (a) End-effector's position to grasp the desired cable segment at [1.0, 0.5, 0.7]. (b) Forces required during unplugging a USB Type A cable.}
    \end{figure}

For each experiment, the robot first flies to around $0.5m$ distance from the cable, then moves forward to grasp the cable segment. Figure~\ref{fig:results-uav-end-effector} illustrates how the end-effector's position can reach the target cable point. Table~\ref{tbl:results-uav} shows the viability of the physical interaction with the perceived DLOs in the simulation if at least $13~\unit{mm}$ position error in grasping can be tolerated in the application. The next future step would be to perform the analysis on the real robot.

    \begin{table}[!t]
    \centering
    \caption{Multirotor end-effector position error (in~$\unit{mm}$) for grasping a cable segment. Trials in Gazebo simulator.}
    \label{tbl:results-uav}
    \begin{tabular}{|c|c|c|c|c|c|c|c|c|c|}
    \hline
    \rowcolor[HTML]{EFEFEF} 
    \# of Tests & Max. Error & Mean Error & Std. Dev. \\ \hline
    20 & 12.92 & 7.84 & 2.91 \\ \hline
    \end{tabular}
    \end{table}

\begin{figure}[!t]
    \centering
        \begin{subfigure}[b]{0.50\linewidth}
            \includegraphics[width=\textwidth]{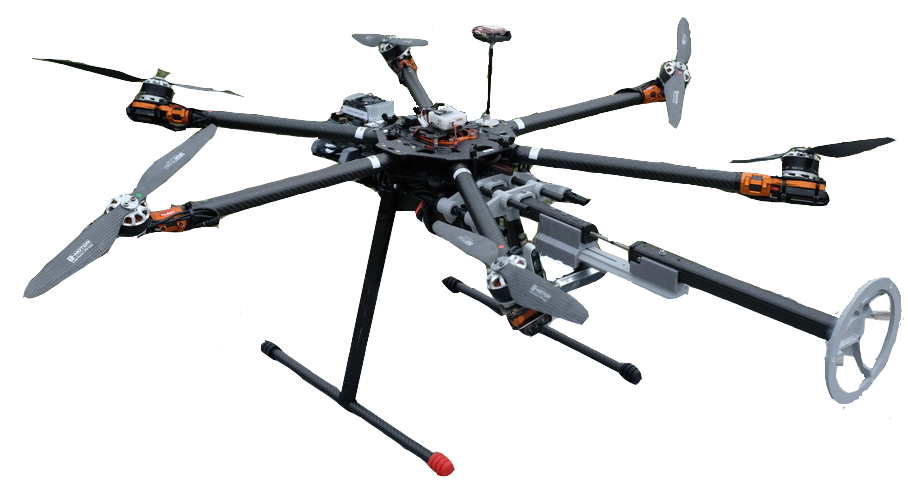}
            \caption{~}
            \label{fig:results-our-uav}
        \end{subfigure}
        \hfill
        \begin{subfigure}[b]{0.20\linewidth}
            \includegraphics[width=\textwidth]{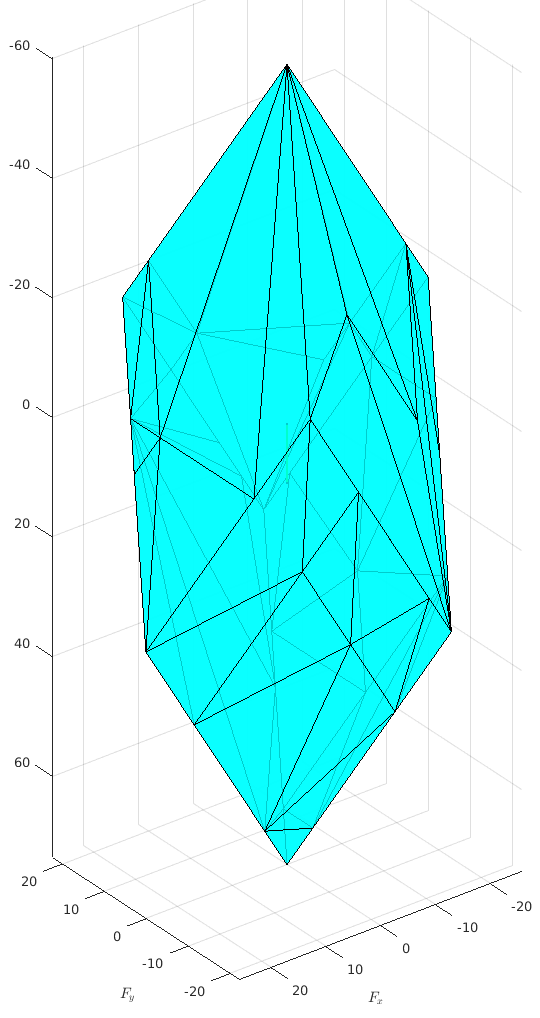}
            \caption{~}
            \label{fig:results-thrust-set}
        \end{subfigure}
        \hfill
        \begin{subfigure}[b]{0.27\linewidth}
            \includegraphics[width=\textwidth]{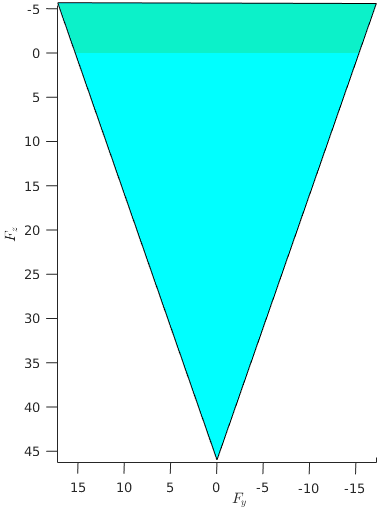}
            \caption{~}
            \label{fig:results-thrust-set-usb-unplug}
        \end{subfigure}
    \caption{Feasibility tests of DLO manipulation forces. (a) Our aerial platform. (b) Force polytopes for our platform. (c) Remaining $y$ and $z$ forces when unplugging a USB cable.}
    \label{fig:results-usb_analysis}
    \end{figure}

On the other hand, aerial robots have limited wrenches compared to ground robots. It is imperative to analyze the feasibility of the physical interaction tasks from the manipulability perspective as well. We computed the force polytopes (Figure~\ref{fig:results-thrust-set}) for our aerial manipulator shown in Figure~\ref{fig:results-our-uav} (see~\cite{keipour2020integration}) and measured the forces required for simple cable-related tasks (such as un/plugging USB cables)~\cite{Keipour-2022-131692, Keipour:2023:unpub:wrench}.

Figure~\ref{fig:results-usb-unplug-plot} shows the example forces measured for unplugging a USB Type A cable which in this scenario is $15.84~\unit{N}$ at the peak. Figure~\ref{fig:results-thrust-set-usb-unplug} shows the available forces for our robot when it is pulling the plug with $15.84~\unit{N}$ directly in its backward direction. The green region shows the remaining forces that allow the robot to keep its altitude. This analysis shows that our aerial robot would be able to unplug the cable in this case, but it is very close to its limits and may not be able to perform a more demanding task.

\section{Future Work} \label{sec:conclusion}

The proposed DLO detection method is a good stepping stone to having a more comprehensive and powerful detection method, and several improvements can be made to the current version. Here are the potential future work directions for both the detection method and its applications:

\begin{itemize}[leftmargin=*]
    \item Detecting which DLO segment is on top of the other at the crossings.
    \item Combining detection with a powerful DLO segmentation and a tracking method to get the full perception system.
    \item Implementing the method more optimally for real-time applications.
    \item Extending and testing the detection method for 3-D.
    \item Performing the feasibility analysis on the real aerial robot.
    \item Test the detection method in a real-world aerial manipulation task.
\end{itemize}

\addtolength{\textheight}{-4.0cm}   % This command serves to balance the column lengths

%%%%%%%%%%%%%%%%%%%%%%%%%%%%%%%%%%%%%%%%%%%%%%%%%%%%%%%%%%%%%%%%%%%%%%%%%%%%%%%%

%%%%%%%%%%%%%%%%%%%%%%%%%%%%%%%%%%%%%%%%%%%%%%%%%%%%%%%%%%%%%%%%%%%%%%%%%%%%%%%%

%%%%%%%%%%%%%%%%%%%%%%%%%%%%%%%%%%%%%%%%%%%%%%%%%%%%%%%%%%%%%%%%%%%%%%%%%%%%%%%%
% \section*{Acknowledgment}

%%%%%%%%%%%%%%%%%%%%%%%%%%%%%%%%%%%%%%%%%%%%%%%%%%%%%%%%%%%%%%%%%%%%%%%%%%%%%%%%

\bibliographystyle{IEEEtran}
%\addbibresource{paper-citations.bib}
\bibliography{paper-citations.bib}

\begin{thebibliography}{10}
\providecommand{\url}[1]{#1}
\csname url@rmstyle\endcsname
\providecommand{\newblock}{\relax}
\providecommand{\bibinfo}[2]{#2}
\providecommand\BIBentrySTDinterwordspacing{\spaceskip=0pt\relax}
\providecommand\BIBentryALTinterwordstretchfactor{4}
\providecommand\BIBentryALTinterwordspacing{\spaceskip=\fontdimen2\font plus
\BIBentryALTinterwordstretchfactor\fontdimen3\font minus
  \fontdimen4\font\relax}
\providecommand\BIBforeignlanguage[2]{{%
\expandafter\ifx\csname l@#1\endcsname\relax
\typeout{** WARNING: IEEEtran.bst: No hyphenation pattern has been}%
\typeout{** loaded for the language `#1'. Using the pattern for}%
\typeout{** the default language instead.}%
\else
\language=\csname l@#1\endcsname
\fi
#2}}

\bibitem{KRISSIAN2000130}
K.~Krissian, G.~Malandain, N.~Ayache, R.~Vaillant, and Y.~Trousset,
  ``Model-based detection of tubular structures in 3d images,'' \emph{Computer
  Vision and Image Understanding}, vol.~80, no.~2, pp. 130--171, 2000.

\bibitem{28493}
T.~Hachaj and M.~R. Ogiela, ``Segmentation and visualization of tubular
  structures in computed tomography angiography,'' in \emph{Intelligent
  Information and Database Systems}, 2012, pp. 495--503.

\bibitem{Wang2021}
B.~Wang, H.~Shi, E.~Cui, H.~Zhao, D.~Yang, J.~Zhu, and S.~Dou, ``A robust and
  efficient framework for tubular structure segmentation in chest ct images,''
  \emph{Technology and Health Care}, vol.~29, pp. 655--665, 2021, 4.

\bibitem{Noble2011}
J.~H. Noble and B.~M. Dawant, ``\BIBforeignlanguage{eng}{A new approach for
  tubular structure modeling and segmentation using graph-based techniques},''
  \emph{\BIBforeignlanguage{eng}{Medical image computing and computer-assisted
  intervention (MICCAI)}}, vol.~14, no. Pt 3, pp. 305--312, 2011.

\bibitem{32226}
C.~Wang, Y.~Hayashi, M.~Oda, H.~Itoh, T.~Kitasaka, A.~F. Frangi, and K.~Mori,
  ``Tubular structure segmentation using spatial fully connected network with
  radial distance loss for 3d medical images,'' in \emph{Medical Image
  Computing and Computer Assisted Intervention (MICCAI)}, 2019, pp. 348--356.

\bibitem{5432191}
A.~Myronenko and X.~Song, ``Point set registration: Coherent point drift,''
  \emph{IEEE Transactions on Pattern Analysis and Machine Intelligence},
  vol.~32, no.~12, pp. 2262--2275, 2010.

\bibitem{5980431}
S.~Javdani, S.~Tandon, J.~Tang, J.~F. O'Brien, and P.~Abbeel, ``Modeling and
  perception of deformable one-dimensional objects,'' in \emph{2011 IEEE
  International Conference on Robotics and Automation}, 2011, pp. 1607--1614.

\bibitem{15711}
O.~Pauly, H.~Heibel, and N.~Navab, ``A machine learning approach for deformable
  guide-wire tracking in fluoroscopic sequences,'' in \emph{Medical Image
  Computing and Computer-Assisted Intervention (MICCAI)}, 2010, pp. 343--350.

\bibitem{wang2020tracking}
Y.~Wang, D.~McConachie, and D.~Berenson, ``Tracking partially-occluded
  deformable objects while enforcing geometric constraints,'' in \emph{2021
  International Conference on Robotics and Automation (ICRA)}, 2021, pp. 1--7.

\bibitem{rastegarpanah}
A.~Rastegarpanah, R.~Howard, and R.~Stolkin, ``Tracking linear deformable
  objects using slicing method,'' \emph{Robotica}, p. 1–19, 2021.

\bibitem{6630714}
J.~Schulman, A.~Lee, J.~Ho, and P.~Abbeel, ``Tracking deformable objects with
  point clouds,'' in \emph{2013 IEEE International Conference on Robotics and
  Automation}, 2013, pp. 1130--1137.

\bibitem{8560497}
Y.~Lai, J.~Poon, G.~Paul, H.~Han, and T.~Matsubara, ``Probabilistic pose
  estimation of deformable linear objects,'' in \emph{International Conference
  on Automation Science and Engineering (CASE)}, 2018, pp. 471--476.

\bibitem{af66d77e53304e6bbb64049bb46193cb}
N.~Padoy and G.~Hager, ``\BIBforeignlanguage{English (US)}{Deformable tracking
  of textured curvilinear objects},'' in \emph{\BIBforeignlanguage{English
  (US)}{2012 23rd British Machine Vision Conference, BMVC 2012}}, 2012.

\bibitem{8206058}
T.~Tang, Y.~Fan, H.-C. Lin, and M.~Tomizuka, ``State estimation for deformable
  objects by point registration and dynamic simulation,'' in \emph{2017
  IEEE/RSJ International Conference on Intelligent Robots and Systems (IROS)},
  2017, pp. 2427--2433.

\bibitem{8206190}
R.~Madaan, D.~Maturana, and S.~Scherer, ``Wire detection using synthetic data
  and dilated convolutional networks for unmanned aerial vehicles,'' in
  \emph{2017 IEEE/RSJ International Conference on Intelligent Robots and
  Systems (IROS)}, 2017, pp. 3487--3494.

\bibitem{8577142}
A.~Zormpas, K.~Moirogiorgou, K.~Kalaitzakis, G.~A. Plokamakis,
  P.~Partsinevelos, G.~Giakos, and M.~Zervakis, ``Power transmission lines
  inspection using properly equipped unmanned aerial vehicle (uav),'' in
  \emph{2018 IEEE International Conference on Imaging Systems and Techniques
  (IST)}, 2018, pp. 1--5.

\bibitem{7532456}
G.~Zhou, J.~Yuan, I.-L. Yen, and F.~Bastani, ``Robust real-time uav based power
  line detection and tracking,'' in \emph{2016 IEEE International Conference on
  Image Processing (ICIP)}, 2016, pp. 744--748.

\bibitem{Dai2020}
\BIBentryALTinterwordspacing
Z.~Dai, J.~Yi, Y.~Zhang, B.~Zhou, and L.~He, ``Fast and accurate cable
  detection using {CNN},'' \emph{Applied Intelligence}, vol.~50, no.~12, pp.
  4688--4707, Dec 2020. [Online]. Available:
  \url{https://doi.org/10.1007/s10489-020-01746-9}
\BIBentrySTDinterwordspacing

\bibitem{7279641}
F.~Tian, Y.~Wang, and L.~Zhu, ``Power line recognition and tracking method for
  uavs inspection,'' in \emph{2015 IEEE International Conference on Information
  and Automation}, 2015, pp. 2136--2141.

\bibitem{7181891}
V.~I. Koshelev and D.~N. Kozlov, ``Wire recognition in image within aerial
  inspection application,'' in \emph{2015 4th Mediterranean Conference on
  Embedded Computing (MECO)}, 2015, pp. 159--162.

\bibitem{PAGNANO2013234}
\BIBentryALTinterwordspacing
A.~Pagnano, M.~Höpf, and R.~Teti, ``A roadmap for automated power line
  inspection. maintenance and repair,'' \emph{Procedia CIRP}, vol.~12, pp.
  234--239, 2013, eighth CIRP Conference on Intelligent Computation in
  Manufacturing Engineering. [Online]. Available:
  \url{https://www.sciencedirect.com/science/article/pii/S2212827113006823}
\BIBentrySTDinterwordspacing

\bibitem{6322366}
J.~Zhang, L.~Liu, B.~Wang, X.~Chen, Q.~Wang, and T.~Zheng, ``High speed
  automatic power line detection and tracking for a uav-based inspection,'' in
  \emph{2012 International Conference on Industrial Control and Electronics
  Engineering}, 2012, pp. 266--269.

\bibitem{8972568}
M.~Yan, Y.~Zhu, N.~Jin, and J.~Bohg, ``Self-supervised learning of state
  estimation for manipulating deformable linear objects,'' \emph{IEEE Robotics
  and Automation Letters}, vol.~5, no.~2, pp. 2372--2379, 2020.

\bibitem{keipour2022ral}
A.~Keipour, M.~Bandari, and S.~Schaal, ``Deformable one-dimensional object
  detection for routing and manipulation,'' \emph{IEEE Robotics and Automation
  Letters}, vol.~7, no.~2, pp. 4329--4336, 2022.

\bibitem{10605}
O.~Merveille, H.~Talbot, L.~Najman, and N.~Passat, ``Tubular structure
  filtering by ranking orientation responses of path operators,'' in
  \emph{Computer Vision (ECCV)}, 2014, pp. 203--218.

\bibitem{BFb0056195}
A.~F. Frangi, W.~J. Niessen, K.~L. Vincken, and M.~A. Viergever, ``Multiscale
  vessel enhancement filtering,'' in \emph{Medical Image Computing and
  Computer-Assisted Intervention (MICCAI)}, 1998, pp. 130--137.

\bibitem{72a4e1c53c}
Y.~Wang, X.~Wei, F.~Liu, J.~Chen, Y.~Zhou, W.~Shen, E.~Fishman, and A.~Yuille,
  ``\BIBforeignlanguage{English (US)}{Deep distance transform for tubular
  structure segmentation in {CT} scans},'' \emph{\BIBforeignlanguage{English
  (US)}{Proceedings of the IEEE Computer Society Conference on Computer Vision
  and Pattern Recognition}}, pp. 3832--3841, 2020.

\bibitem{SAHA20173}
P.~K. Saha, G.~Borgefors, and G.~{Sanniti di Baja}, ``Chapter 1 -
  skeletonization and its applications – a review,'' in
  \emph{Skeletonization: Theory, Methods and Applications}, 2017, pp. 3--42.

\bibitem{keipour2013omnifont}
A.~Keipour, M.~Eshghi, S.~M. Ghadikolaei, N.~Mohammadi, and S.~Ensafi,
  ``Omnifont {Persian OCR} system using primitives,'' \emph{arXiv:2202.06371},
  pp. 1--5, 2013.

\bibitem{1164959}
P.~Maragos and R.~Schafer, ``Morphological skeleton representation and coding
  of binary images,'' \emph{IEEE Transactions on Acoustics, Speech, and Signal
  Processing}, vol.~34, no.~5, pp. 1228--1244, 1986.

\bibitem{Gong2018}
X.-Y. Gong, H.~Su, D.~Xu, Z.-T. Zhang, F.~Shen, and H.-B. Yang, ``An overview
  of contour detection approaches,'' \emph{International Journal of Automation
  and Computing}, vol.~15, no.~6, pp. 656--672, Dec 2018.

\bibitem{keipour2021ral}
A.~Keipour, G.~A. Pereira, and S.~Scherer, ``Real-time ellipse detection for
  robotics applications,'' \emph{IEEE Robotics and Automation Letters}, vol.~6,
  no.~4, pp. 7009--7016, 2021.

\bibitem{suzuki1985topological}
S.~Suzuki and K.~Abe, ``Topological structural analysis of digitized binary
  images by border following,'' \emph{Computer vision, graphics, and image
  processing}, vol.~30, no.~1, pp. 32--46, 1985.

\bibitem{keipour2022iros}
A.~Keipour, M.~Bandari, and S.~Schaal, ``Efficient spatial representation and
  routing of deformable one-dimensional objects for manipulation,''
  \emph{arXiv:2202.06172}, pp. 1--7, 2022.

\bibitem{keipour2020integration}
A.~Keipour, M.~Mousaei, A.~T. Ashley, and S.~Scherer, ``Integration of
  fully-actuated multirotors into real-world applications,'' \emph{arXiv
  preprint arXiv:2011.06666}, 2020.

\bibitem{Keipour-2022-131692}
A.~Keipour, ``Physical interaction and manipulation of the environment using
  aerial robots,'' Ph.D. dissertation, Carnegie Mellon University, Pittsburgh,
  PA, May 2022.

\bibitem{Keipour:2023:scitech:simulator}
\BIBentryALTinterwordspacing
A.~Keipour, M.~Mousaei, D.~Bai, J.~Geng, and S.~Scherer, ``{UAS} simulator for
  modeling, analysis and control in free flight and physical interaction,'' in
  \emph{AIAA SciTech 2023 Forum}.\hskip 1em plus 0.5em minus 0.4em\relax
  American Institute of Aeronautics and Astronautics, Jan 2023. [Online].
  Available: \url{https://arc.aiaa.org/doi/abs/10.2514/6.2023-1279}
\BIBentrySTDinterwordspacing

\bibitem{Keipour:2023:icra-workshop:simulator}
A.~Keipour, M.~Mousaei, and S.~Scherer, ``A simulator for fully-actuated
  uavs,'' in \emph{Workshop on The Role of Robotics Simulators for Unmanned
  Aerial Vehicles, 2023 IEEE International Conference on Robotics and
  Automation (ICRA)}, Jun 2023, pp. 1--4.

\bibitem{Keipour:2023:unpub:wrench}
A.~Keipour, M.~Mousaei, J.~Geng, and S.~Scherer, ``Real-time wrench-set
  analysis and applications for aerial robots,'' Aug 2023, in press.

\end{thebibliography}

\end{document}